\documentclass[preprint,12pt]{elsarticle}

\usepackage{amssymb}
\usepackage{amsmath}

\usepackage[table]{xcolor}
\usepackage{setspace}
\usepackage{pdflscape}
\usepackage{amsthm}
\usepackage{makecell}

\theoremstyle{definition}

\usepackage{url}

\usepackage{makecell}
\usepackage{amsmath}
\usepackage{multicol}
\usepackage{multirow}

\journal{Natural Language Processing Journal}

\begin{document}

\begin{frontmatter}



\title{Comparison of Open-Source and Proprietary LLMs for Machine Reading Comprehension: A Practical Analysis for Industrial Applications}


\affiliation[a]{
    organization={Novelis Research and Innovation Lab}, 
    address={40 av. des Terroirs de France},
    city={Paris},
    postcode={75012},
    country={France}
}

\author[a]{Mahaman Sanoussi Yahaya Alassan}
\author[a]{Jessica López Espejel}
\author[a]{Merieme Bouhandi}
\author[a]{Walid Dahhane}
\author[a]{El Hassane Ettifouri}

\begin{abstract}
Large Language Models (LLMs) have recently demonstrated remarkable performance in various Natural Language Processing (NLP) applications, such as sentiment analysis, content generation, and personalized recommendations. Despite their impressive capabilities, there remains a significant need for systematic studies concerning the practical application of LLMs in industrial settings, as well as the specific requirements and challenges related to their deployment in these contexts. This need is particularly critical for Machine Reading Comprehension (MCR), where factual, concise, and accurate responses are required. To date, most MCR rely on Small Language Models (SLMs) or Recurrent Neural Networks (RNNs) such as Long Short-Term Memory (LSTM). This trend is evident in the SQuAD2.0 rankings on the Papers with Code table. This article presents a comparative analysis between open-source LLMs and proprietary models on this task, aiming to identify light and  open-source alternatives that offer comparable performance to proprietary models.
\end{abstract}

\begin{graphicalabstract}
\resizebox{\linewidth}{!}{
\includegraphics{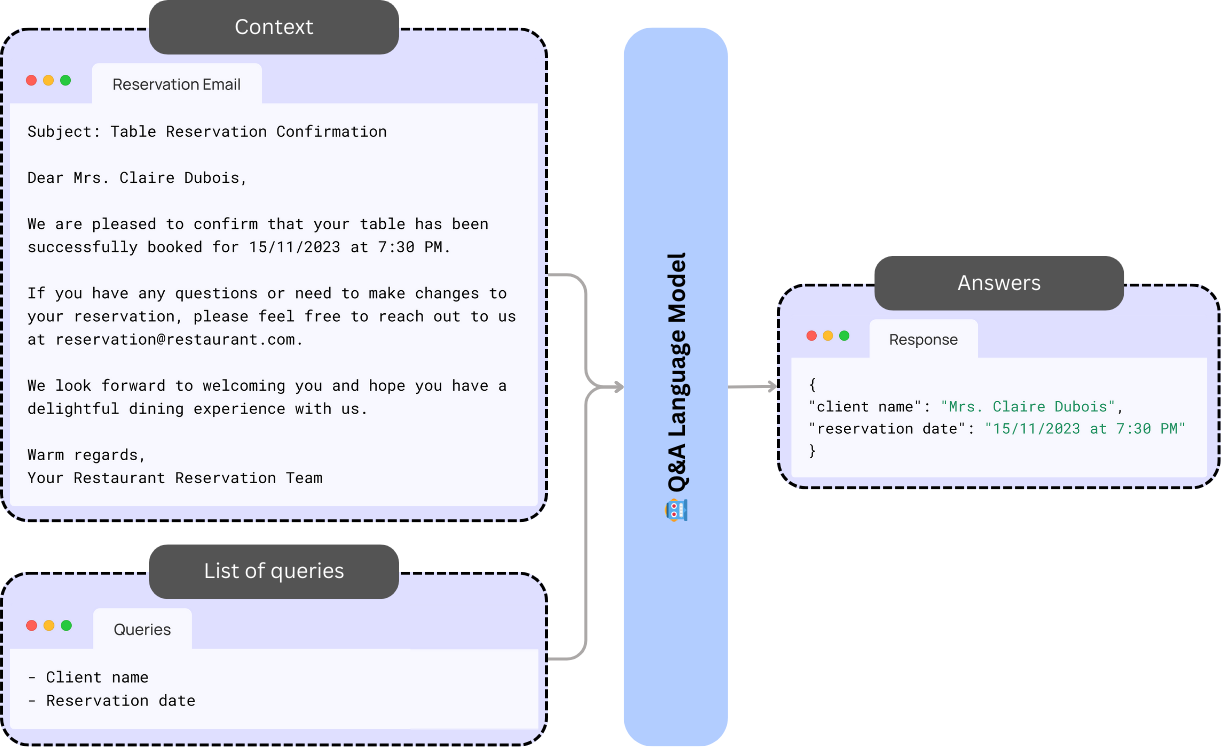}
}
\end{graphicalabstract}

\begin{highlights}
    \item Comprehensive comparative analysis of open-source and proprietary LLMs for Machine Reading MRC in industrial contexts.
    \item Focus on identifying resource-efficient open-source alternatives to proprietary models for practical deployment.
    \item GPT-3.5 and GPT-4 excelled with exact match scores of 83\% and 87\% and ROUGE-2 scores of 91\% and 83\%, respectively.
    \item Mistral models, with Mistral-7b-openorca, achieved comparable results to GPT-3.5, with an exact match score of 83\% and ROUGE-2 score of 80\%.
    \item Phi models, with Dolphin-2.6-phi-2, delivered the fastest inference time (25.72 ms), making it suitable for real-time applications, though accuracy was lower (70\% exact match).
    \item LLaMA Models, with Llama-2-7b-chat, provide solid MCR performance with low memory usage, making them ideal for resource-constrained environments.
\end{highlights}

\begin{keyword}
Natural Language Processing \sep Large Language Models \sep Machine Reading Comprehension   \sep Low-Cost models
\end{keyword}

\end{frontmatter}



\newpage

\section{Introduction}
\label{sec: introduction}

    Natural Language Processing (NLP) has been revolutionized by the emergence of Large Language Models (LLMs), enabling advancements in tasks such as machine translation~\citep{zhao2023transformer,bo2024_attentionMT, huang2024moceadaptivemixturecontextualization}, text summarization~\citep{li2023improving,doan2023too}, and knowledge extraction~\citep{gu2023distilling,xu2024take}. These models leverage large and sophisticated architectures to generate coherent and contextually relevant outputs, significantly expanding the scope of natural language understanding and generation. Prominent examples include GPT-4~\citep{chatGPT4}, Claude 3 Opus~\citep{claude}, and Mixtral 8x7B~\citep{jiang2024_mixtral}. While LLMs have reshaped the language technology landscape, their practical applications in industrial contexts remain an area of active exploration. This study focuses on Machine Reading Comprehension (MRC), an NLP task where LLMs demonstrate the ability to interpret and answer questions based on textual content. MRC systems are particularly valuable for industries that depend on document analysis, automated customer support, and knowledge management systems, making them a key component of modern AI-driven solutions.

    While LLMs have great potential, their adoption in real-world industrial scenarios involves complex trade-offs. Proprietary LLMs often deliver state-of-the-art performance on benchmark datasets, but their substantial computational demands, high costs, and limited transparency pose challenges for widespread adoption. On the other hand, open-source LLMs offer benefits such as lower costs and customizable deployments, yet their performance may fall short in resource-intensive applications. These trade-offs are especially pronounced in MRC tasks, where industries require solutions that balance accuracy, inference speed, and resource usage under operational constraints.

    In this context, the current study addresses the gap in systematic evaluations of LLMs for industrial MRC applications. By comparing open-source and proprietary LLMs on metrics relevant to real-world deployments, this work aims to guide stakeholders in selecting models that align with their specific needs. The contributions of this study are threefold: (1) it benchmarks open-source and proprietary LLMs across performance metrics such as accuracy, efficiency, and scalability; (2) it identifies practical trade-offs for deploying these models in resource-constrained environments; and (3) it highlights the potential of high-performing open-source models as viable alternatives to proprietary systems. This analysis aims to provide actionable insights for industries seeking to integrate MRC capabilities into their workflows while navigating operational challenges.
    
\section{Background and Related Work}
\label{sec:related-work}

    \subsection{Overview of LLMs}
   LLMs have become a cornerstone of modern NLP, transforming the field with their exceptional ability to generate, comprehend, and manipulate human language with unprecedented accuracy. These models are increasingly integrated into industrial applications such as customer service automation~\citep{soni2023large, shi2024chops, pandya2023automating}, virtual assistants~\citep{kernan2023harnessing, luo2023valley, ross2023programmer}, and information retrieval systems~\citep{liu2023reta, hou2024large, agrawal2022large}, among others. These models are typically based on deep learning architectures, particularly Transformer models~\citep{vaswani2017attention}, which are highly suited for text generation.

   Several studies have explored the performance of open-source and proprietary LLMs~\citep{yu2023open,Lopez2025_LowCost}, providing valuable insights into their strengths and limitations. Open-source models, such as Llama 2~\citep{touvron2023_llama2}, have demonstrated that they can rival proprietary models like GPT-3.5~\citep{chatGPT3.5} when fine-tuned, highlighting that task-specific model selection and prompt engineering are crucial for achieving good results~\citep{yu2023open}. While proprietary models like GPT-4 dominate certain areas such as misinformation detection, open-source models like Zephyr-7B~\citep{tunstall2023_zephyr} offer competitive and more accessible alternatives, with the added benefit of being more cost-effective~\citep{vergho2024comparing}. Additionally, open-source models like Mixtral 8x7B~\citep{jiang2024_mixtral} have excelled in few-shot learning tasks, such as retrieval-augmented generation in BioASQ \footnote{\url{https://www.bioasq.org/}}, showcasing their adaptability and efficient use of resources~\citep{ateia2024can}. Despite continuous improvements in GPT models, they still struggle with complex knowledge-based question answering~\citep{tan2023can}, where traditional methods often outperform them, although techniques like chain-of-thought prompting~\citep{Wei2022_COT} have improved their performance.

   The growing use of LLMs in domain-specific applications such as healthcare, education, and finance has raised questions about the effectiveness of both open-source and proprietary models. In healthcare, open-source models like Llama 3~\citep{grattafiori2024llama3herdmodels} have shown comparable performance to proprietary models like ChatGPT in radiology exam questions, underscoring the advantages of open-source models in terms of privacy and customization for medical applications~\citep{adams2024llama}. Furthermore, Llama2-70B demonstrated strong performance in radiology report classification~\citep{dorfner2024open}, making it a promising candidate for healthcare applications. In education, open-source Learning Management Systems (LMS) outperform proprietary systems in flexibility, cost, and adaptability, particularly in higher education, offering a more affordable and effective alternative~\citep{pankaja2013proprietary}. Similarly, in finance, open-source models like LLaMA have shown comparable performance to proprietary models like ChatGPT in financial document analysis~\citep{panwar2023comparative}, further solidifying the value of open-source models in industry-specific applications.

   Ongoing efforts to improve open-source models have been the focus of several studies, aiming to close the performance gap that exists with proprietary systems. Iterative self-refinement has led to the significant improvement of models like Vicuna-7B~\citep{vicuna2023}, reducing this performance gap and contributing to the democratization of AI by making powerful models more accessible~\citep{shashidhar2023democratizing}. Fine-tuning has also proven effective in enhancing the capabilities of open-source models for specialized tasks, such as medical evidence summarization, where fine-tuned models demonstrate competitive performance compared to proprietary counterparts~\citep{zhang2024closing}. However, challenges remain when attempting to fine-tune weaker models to imitate stronger proprietary LLMs. While promising, this approach exposes performance gaps, particularly in areas such as coding and problem-solving, highlighting the need for more robust base models~\citep{gudibande2023false}.

   Despite these successes, LLMs still face several challenges that hinder their overall performance. The rapid growth of the field has emphasized the need for standardized benchmarks to assess issues such as licensing, bias, and robustness, especially for open-source models~\citep{schur2023comparative}. Furthermore, LLMs still struggle with complex question-answering (QA) tasks that require hybrid evidence integration or handle ambiguous queries. Advancements in retrieval and reasoning mechanisms are crucial to addressing these challenges~\citep{prabhu2024dexter}. While LLMs excel in conversational QA tasks, improvements are needed to enhance the consistency and specificity of their responses~\citep{rangapur2024battle}. Additionally, performance gaps in specialized tasks, such as data privacy policy analysis, underscore the need for optimized AI-human collaboration and advanced infrastructure to improve LLM capabilities~\citep{filipovska2024benchmarking}.

    While numerous studies have explored the performance of open-source and proprietary LLMs in various question-answering tasks, there is a notable gap in research specifically focused on comparing these models for Machine Reading MRC. For example, ~\cite{yagnik2024medlm} discusses progress in medical Q\&A, which is closely related to MRC, but it specifically refers to closed-book medical Q\&A. Similarly, ~\cite{tan2023can} mentions how GPT models improve with versions but still underperform in knowledge-based complex question-answering. The lack of direct comparisons between LLMs for MRC tasks motivated this study.

    \subsection{MCR Tasks}
    
    Technological advancements have led to the omnipresence of textual data in the society. From SMS messages on mobile phones to various file formats such as pdf, docx, and ppt, to micro-blogs like X\footnote{\url{https://twitter.com/}} and social media comments on platforms like Facebook or Google+, textual data permeate our daily lives. However, the sheer abundance of textual information exacerbates the challenge of information overload. Managing this information overload has become a formidable challenge that information systems must address~\citep{dietzmann2022artificial}. This necessitates the development of tools to automate information retrieval processes, facilitate access to information, and alleviate information overload.
    
    QA is defined as a multidisciplinary research area that intersects Information Retrieval (IR), Information Extraction (IE), and NLP \citep{allam2012question}. Its goal is to provide accurate answers to user queries by extracting relevant information from textual documents or databases. Unlike traditional search engines that return ranked lists of documents, QA systems deliver direct and specific responses tailored to user questions. An essential process at the heart of QA systems is MRC, which plays a crucial role in extracting precise answers from text. In QA pipelines, MRC typically operates in the final stage, following the information retrieval phase. The MRC model then focuses on understanding these passages and extracting the most appropriate answer to the question posed. By mimicking human reading and comprehension, MRC is indispensable for enhancing the accuracy and relevance of answers. MRC can be formulated as follows: given a context (a given document) $c$ and a question $q$, the objective is to have a function $f$, which returns the appropriate answer $a$ given the context $c$ and related to the question $q$ such as $f(c,q) = a$.

   From another hand, \citet{baradaran2022survey} segmented MRC approaches into three main categories: rule-based methods, classical machine learning methods, and deep learning methods. Rule-based approaches rely on domain-specific rules, often crafted by linguists, to match relevant text passages with questions. Although effective in targeted contexts, these methods are limited by rule incompleteness and require extensive adaptation for each application domain. Classical machine learning methods, while avoiding explicit rules, rely on manually engineered features to train models capable of mapping questions to relevant text spans. This approach requires significant expertise in feature engineering and remains sensitive to the specific domain. Deep learning methods have advanced MRC significantly by allowing models to autonomously learn complex text representations. Using architectures such as Transformer-based networks, these methods excel at contextualizing responses and adapting to a wide range of question types.

   MRC process enables QA systems to find wide-ranging applications across various domains. In information retrieval, QA systems excel at extracting relevant data from extensive textual sources such as databases, news articles, and academic publications~\citep{kobayashi2000information, li2024matching}. These systems can address both simple factual questions and complex inquiries requiring in-depth content analysis, as demonstrated by~\cite{qiu2018qa4ie}, who introduced QA4IE, an Information Extraction (IE) framework that utilizes question-answering techniques to handle diverse data relationships. In automated customer support, QA systems play a pivotal role in chatbots, handling frequently asked questions and offering immediate guidance on products or services. They also support decision-making by providing professionals with precise answers to specialized queries, such as recommending treatments in the medical field based on clinical data~\citep{zhang2018medical}. In data analysis, QA systems extract vital information from unstructured datasets, aiding market analysis and business intelligence~\citep{chen2017reading}. Additionally, they enhance online learning platforms by offering instant answers, promoting autonomy, and improving learning efficiency~\citep{hung2005applying}. In legal research, QA systems assist professionals by streamlining research on laws and precedents, enabling informed decision-making~\citep{huang2020aila}.

    \section{Methodology}
        In this section, we outline our approach for evaluating and comparing the performance of proprietary and open-source language models in efficient question-answering for industrial applications. First, we present the baselines and reference models used in our study. Next, we describe our dataset, highlighting its characteristics and its relevance to industrial applications. We also present the evaluation metrics we use to assess model performance. Lastly, we detail our experimental setup, including the data processing pipeline and the specific parameters used to ensure the reproducibility of our results.

    \subsection{Selection of LLMs}
        For our study on Machine Reading Comprehension (MRC), we conducted an in-depth evaluation of both proprietary and open-source language models, selecting a diverse set of high-performing models based on their relevance and efficiency in similar tasks. Proprietary models include GPT-3.5~\citep{chatGPT3.5} and GPT-4~\citep{chatGPT4}, known for their advanced architectures and robust performance in complex tasks. In contrast, open-source models such as Mistral-7B-Instruct-v0.2~\citep{jiang2023mistral}, Llama 2 7B Chat~\citep{touvron2023_llama2}, Dolphin-2\_6-Phi-2~\citep{dolphin_phi}, Dolphin-2.6-Mistral-7B~\citep{dolphin_mistral}, and Mistral-7B-OpenOrca~\citep{mukherjee2023_orca} offer greater flexibility, allowing for fine-tuning and potential cost advantages.

        The selection criteria for these models include their architectural features and task-specific optimizations. We intentionally focused on open-source models with 7 billion parameters, which offer a favorable balance between computational efficiency and performance, making them ideal for production use. Larger open-source models were excluded due to their higher computational costs.

        \textbf{Architectural features}: GPT-3.5~\citep{chatGPT3.5} and GPT-4~\citep{chatGPT4} utilize autoregressive transformer architectures, with enhancements such as Reinforcement Learning from Human Feedback (RLHF) to improve multitask dialogue and reasoning capabilities. Open-source models like Mistral-7B-Instruct-v0.2 and Llama 2 7B Chat adopt innovative attention mechanisms to optimize performance. For instance, Mistral-7B incorporates techniques for efficient processing of long sequences, while Llama 2 7B Chat is fine-tuned to align with human preferences, balancing helpfulness and safety.

        \textbf{Fine-tuning and task-specific optimization}: Proprietary models like GPT-3.5 and GPT-4 are designed for general-purpose use, performing well across a wide range of tasks without requiring specific fine-tuning. As such, they serve as strong benchmarks against which to measure other models, although further task-specific fine-tuning may be needed for specialized applications. Selected open-source models are primarily optimized for instruction-following, which improves their responsiveness to context-specific prompts. Mistral-7B-Instruct-v0.2 and Mistral-7B-OpenOrca, for instance, are instruction-tuned, with OpenOrca outperforming other 7B models and even some 13B models on the Hugging Face leaderboard. Dolphin-2\_6-Phi-2 and Dolphin-2.6-Mistral-7B are uncensored models \footnote{\url{https://erichartford.com/uncensored-models}} designed to enable diverse, customizable alignments that cater to various cultural, ideological, and legal needs, promoting freedom of choice, intellectual curiosity, and user control, while supporting the development of composable alignment frameworks.
    
     \subsection{Dataset}
        The dataset used in this study was generated using OpenAI's ChatGPT-3.5 model~\citep{chatGPT3.5}. The data generation process involved applying the few-shot prompting technique, followed by instructing the model to generate unstructured text examples from various domains, such as banking, healthcare, e-commerce, and others. The dataset is structured in JSON format. It consists of 40 diverse samples, each containing the following components:
    
    \begin{itemize}
        \item
        Context: The raw unstructured text which may be an HTML string, an excerpt from an email, or text extracted from a PDF document, among other types.
        \item
        Request: A series of queries designed to simulate potential user inquiries related to the given text.
        \item
        Output: The expected result set is presented in a structured JSON format, serving as a benchmark for evaluating model performance in extracting relevant information in response to the provided queries.
    \end{itemize}

    This dataset has been validated to ensure the quality of the generated examples, ensuring they meet the criteria for diversity and representativeness of real-world scenarios encountered in the targeted application domains. Although the examples were generated by ChatGPT-3.5, a manual review was performed to ensure their relevance and accuracy in the context of the evaluation. This validation is particularly important for more complex samples, where answers are not explicitly present in the context.    
    
    The dataset is further categorized into three levels of difficulty:
        \begin{itemize}
        \item 
        Easy: Answers to expected questions are explicit within the context. Table~\ref{tab:easy_example} shows an example.

        \begin{table}[ht!]
            \centering
            \small
            \begin{tabular}{|p{1.0\linewidth}|}
                \hline
                \texttt{"Context": "Patient: John Doe, SSN 123-45-6789, had a consultation on 10/10/2023 with Dr. Smith. For billing inquiries, email billing@medicenter.com."} \\
                \texttt{}\\
                \texttt{"Request": [}\\
                \texttt{\hspace*{6em}"What is the patient name?",}\\
                \texttt{\hspace*{6em}"What is the social Security Number?",}\\
                \texttt{\hspace*{6em}"What is the consultation date?",}\\
                \texttt{\hspace*{6em}"What is the billing inquiries email?"}\\
                \texttt{]},\\
                \texttt{}\\
                \texttt{"Output": \{}\\
                \texttt{\hspace*{1em}"What is the patient name?": "John Doe",}\\
                \texttt{\hspace*{1em}"What is the social Security Number?": "123-45-6789",}\\
                \texttt{\hspace*{1em}"What is the consultation date?": "10/10/2023",}\\
                \texttt{\hspace*{1em}"What is the billing inquiries email?": "billing@medicenter.com"}\\
                \texttt{\}},\\
                \hline
            \end{tabular}
            \caption{Easy-level difficulty sample}
            \label{tab:easy_example}
        \end{table}

        \item 
        Medium: Two questions may have answers of the same type. Table~\ref{tab:medium_example} shows an example.
    
        \begin{table}[ht!]
            \centering
            \small
            \begin{tabular}{|p{1.0\linewidth}|}
                \hline
                \texttt{"Context": "Mr. Jacques Durand, your order \#12345 placed on 20/10/2023 will be delivered on 22/10/2023. For tracking, contact support@shop.com."} \\
                \texttt{}\\
                \texttt{"Request": [}\\
                \texttt{\hspace*{6em}"What is the client name?",}\\
                \texttt{\hspace*{6em}"What is the order date?",}\\
                \texttt{\hspace*{6em}"What is the delivery date?",}\\
                \texttt{\hspace*{6em}"What is the support email?"}\\
                \texttt{]},\\
                \texttt{}\\
                \texttt{"Output": \{}\\
                \texttt{\hspace*{6em}"What is the client name?": "Mr. Jacques Durand",}\\
                \texttt{\hspace*{6em}"What is the order date?": "20/10/2023",}\\
                \texttt{\hspace*{6em}"What is the delivery date?": "22/10/2023"}\\
                \texttt{\}},\\
                \hline
            \end{tabular}
            \caption{Medium-level difficulty sample}
            \label{tab:medium_example}
        \end{table}

        \item 
        Complex: Answers to expected questions are not explicit within the context. An example is shown in Table~\ref{tab:complex_example}.
    
        \begin{table}[ht!]
            \centering
            \small
            \begin{tabular}{|p{1.0\linewidth}|}
                \hline
                \texttt{"Context": "Dear BankXYZ Support, I recently noticed a transaction on my account that I did not authorize. It's listed as 'Online Purchase \$150' on 12th March. I'd like this to be investigated as soon as possible. Regards, John Appleseed"} \\
                \texttt{}\\
                \texttt{"Request": [}\\
                \texttt{\hspace*{6em}"What is the type of issue?",}\\
                \texttt{\hspace*{6em}"What is the transaction detail?",}\\
                \texttt{\hspace*{6em}"What is the transaction date?"}\\
                \texttt{]},\\
                \texttt{}\\
                \texttt{"Output": \{}\\
                \texttt{\hspace*{2em}"What is the type of issue?": "Unauthorized transaction",}\\
                \texttt{\hspace*{2em}"What is the transaction detail?": "Online Purchase \$150",}\\
                \texttt{\hspace*{2em}"What is the transaction date?": "12th March"}\\
                \texttt{\}},\\
                \hline
            \end{tabular}
            \caption{Complex-level difficulty sample}
            \label{tab:complex_example}
        \end{table}
        
    \end{itemize}

    While the dataset contains only 40 samples, which may seem limited, it represents a pragmatic approach for exploring information extraction in diverse scenarios. Expansion of the dataset is planned for future iterations, integrating an automated generation and validation process through additional tools to increase the diversity of prompts and the quality of examples.

    
    \subsection{Experimental setup}
    
\begin{figure}[ht!]
        	\begin{center}
            \includegraphics[width=0.9\textwidth]{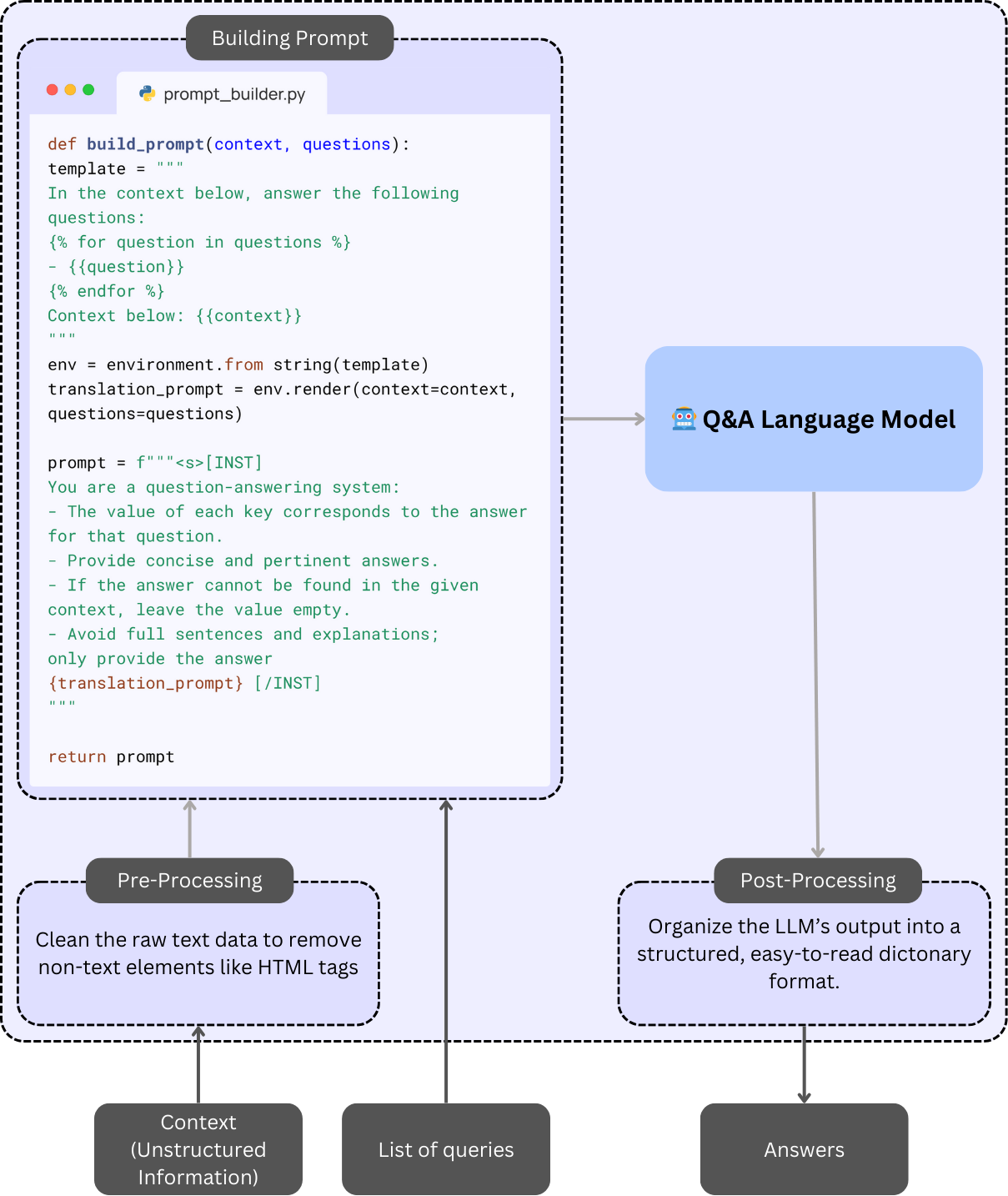}
        	\end{center}
            \vspace{-1em}
        	\caption{Overview of our pipeline. The pre-processing stage cleans raw text data, removing non-text elements. The prompting stage builds dynamic prompts with the given user query and document (namely, the context of the query). The post-processing stage formats the LLM's responses and structures them into a dictionary.}
             \vspace{-1em}
        	\label{fig:python_generation}
        \end{figure}

    We use a pipeline (Figure~\ref{fig:python_generation}) that centers on harnessing the capabilities of LLMs to extract structured information from unstructured text contexts based on a provided list of queries. A query can be a question or an information to extract.

    The pipeline consists of the following preprocessing and postprocessing steps, ensuring that the data is well-prepared for input and subsequent integration into applications:

    \begin{itemize}
        \item
        \textbf{Preprocessing}: In light of the prevalence of web-derived content in contemporary data, ensuring that our raw text is devoid of non-textual or extraneous elements, particularly HTML tags, is crucial. To achieve this, we employ the BeautifulSoup Python library~\footnote{\url{https://tedboy.github.io/bs4_doc/}}, effectively cleaning our text and removing any hindrances.
        \item
        \textbf{Prompting}: The creation of prompts involves integrating the context with user-defined queries. This prompt serves as input for LLMs, guiding them in generating responses that align with the given queries. 
        \item
        \textbf{Postprocessing}: This critical step ensures continuity and coherence in the extracted information by presenting it in a structured format. Specifically, we structure the extracted information into a dictionary format, where keys represent predefined variables, and the corresponding values denote the extracted information.
    \end{itemize}

    We evaluate various model variants denoted as $model$-$name.Q\_j\_K_Size$, where $model$-$name$ represents the model's name, which can be one of the following: llama-2-7b-chat, mistral-7b-instruct-v0.2, mistral-7b-openorca, or dolphin-2.6-mistral-7b. The term $Q\_j\_K\_Size$ specifies a particular quantization method, where $j$ indicates the bit-width used, $K$ denotes the application of K-means clustering during quantization, and $Size$ corresponds to the size of the model after quantization, with $S$, $M$, and $L$ referring to Small, Medium, and Large, respectively.

      \begin{figure}[ht!]
        	\begin{center}
            \includegraphics[width=1\textwidth]{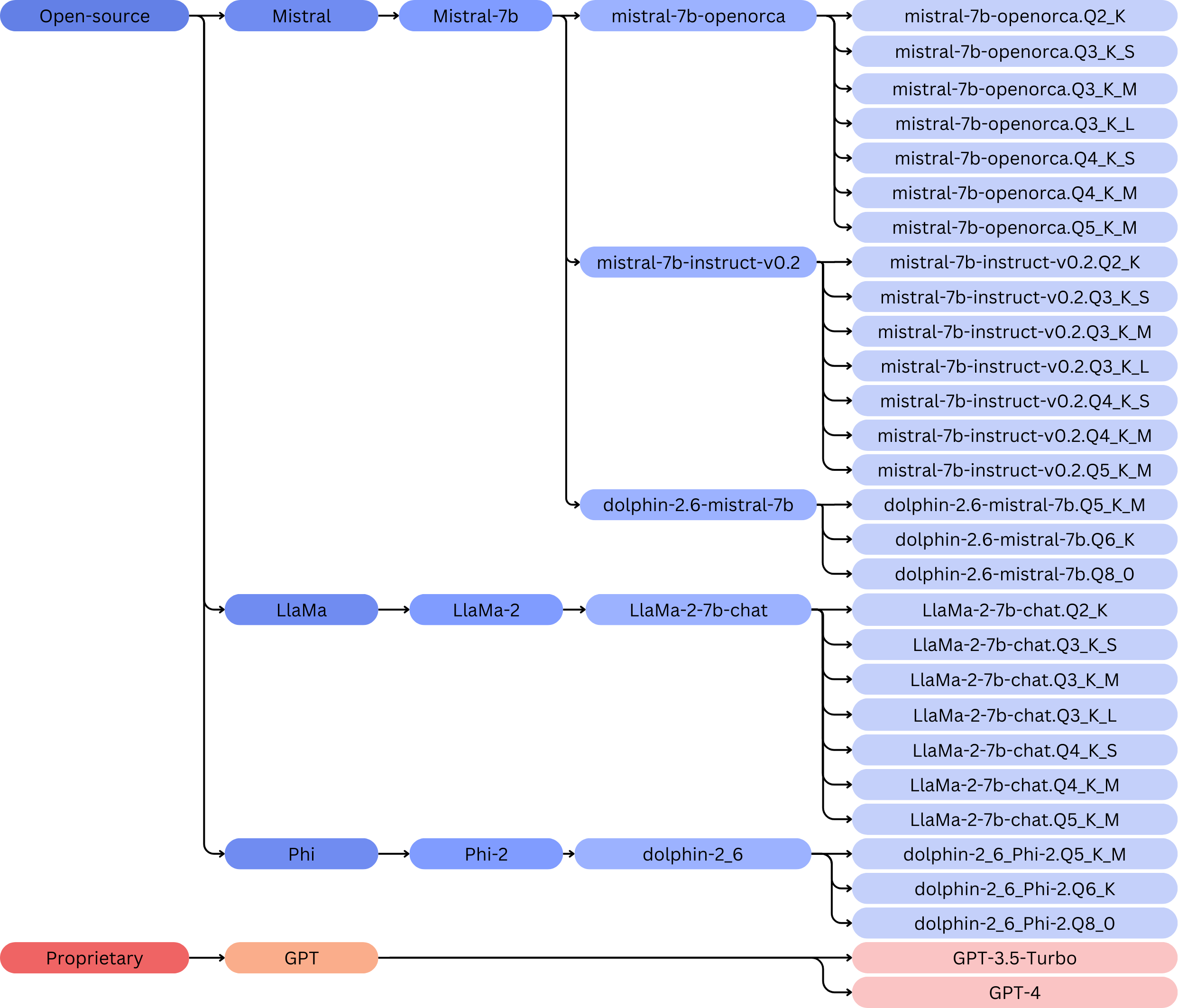}
        	\end{center}
        	\caption{Evaluated Models}
        	\label{fig:evaluated_models}
        \end{figure}

     To conduct our experiments on the CPU, we used a machine equipped with the following hardware: Processor: 11th Gen Intel(R) Core(TM) i7-11850H @ 2.50GHz, RAM: 32.0 GB, Operating System: 64-bit, x64-based processor.

\section{Results and Discussion}
    This section presents an analysis of the performance of various language models in MCR tasks using our dataset, evaluated using exact match and ROUGE-L metrics. These measures provide insights into the models' accuracy and semantic understanding, essential for assessing their suitability for real-world applications. Additionally, the discussion considers inference speed, resource efficiency, and practical applicability, highlighting the strengths, limitations, and trade-offs of each model family

    \textbf{GPT3 models}. The GPT-3 family, comprising GPT-3.5 and GPT-4, demonstrates exceptional performance in machine reading comprehension tasks, with both models achieving high accuracy and robust semantic understanding. GPT-3.5 consistently delivers strong results across various evaluation metrics, excelling in exact match (83\%) and Rouge-L (91\%) scores, even when compared to larger models like GPT-4. The GPT-4 model shows a notable improvement in both semantic reasoning and contextual understanding, achieving the highest Rouge-L score of 94\% on the same tasks. A manual evaluation reveals that these models achieve 99\% accuracy for GPT-3.5 and 100\% for GPT-4 on our dataset, with no significant performance difference observed between the two. In terms of inference speed, GPT-3.5 operates more efficiently, with an inference time of 5.61 ms, whereas GPT-4 requires a longer inference time of 13.21 ms.

    \textbf{Phi2 models}. Our evaluation of Phi-2 models revealed notable insights into their performance across different quantization levels (5, 6, and 8 bits). Among these, the Phi-2.Q5\_K\_M configuration achieved the highest scores, with 70\% exact match (EM) and 87\% ROUGE-L. This result marginally outperformed Phi-2.Q6\_K and Phi-2.Q8\_0, which scored 70\% EM and 85\% ROUGE-L, and 69\% EM and 86\% ROUGE-L, respectively. The results suggest that the finer granularity preserved in the 5-bit quantization offers a slight advantage in capturing nuanced information compared to the higher bit-width configurations. However, despite this achievement, Phi-2 models generally performed at the lower end of the spectrum when compared to other evaluated models.
    
    One key limitation observed with Phi-2 models is their strong dependency on syntax. The models struggled with paraphrased questions, where the wording differed from the original context, and failed to infer answers when the terms in the questions were semantically similar but not explicitly present in the context. This reliance on surface-level matches limits their ability to generalize across diverse formulations of the same query. Additionally, Phi-2 models demonstrated reduced precision when tasked with extracting structured information, such as datetime values. For instance, they were often unable to retrieve both the date and time components accurately, frequently extracting only partial information (e.g., the date without the time).

    \textbf{Mistral models}. The evaluation of the Mistral models—Mistral-7B-Instruct, Mistral-7B-OpenOrca, and Dolphin-2.6-Mistral—highlights their strong performance in machine reading comprehension tasks, with distinct capabilities and trade-offs across different quantization levels. Overall, configurations using Q3 and Q5 quantization consistently deliver better accuracy and robustness.The Mistral-7B-OpenOrca model, in its Q3\_K\_M configuration, stands out as the best performer, achieving 83\% exact match and 90\% ROUGE-L, while maintaining a competitive inference time of 42 ms.

    The Mistral models, as a group, outperform the Phi-2 and Llama families, especially in terms of semantic understanding and precision. They achieve results comparable to larger proprietary models like GPT-3.5 while requiring fewer computational and memory resources. The Mistral models demonstrate a notable ability to handle paraphrased and rephrased questions. However, similar to the Phi-2 models, they struggle with extracting specific types of information, such as full datetime values, where they only retrieve partial information.

    \textbf{Llama models}. Despite not achieving the top results among open-source models, Llama models demonstrate solid capabilities in machine reading comprehension tasks, making them a viable option for resource-constrained applications. The Q3\_K\_S and Q3\_K\_L configurations stand out, achieving 76\% exact match and 88\% ROUGE-L, with a modest inference time of 153 ms and 91 ms, respectively.

    Compared to other open-source models, such as the Mistral and Phi-2 families, the LLaMA-2-7B-Chat models perform moderately well but fall short in specific metrics. For example, while they surpass the Phi-2 models in both exact match and ROUGE-L scores, they do not reach the precision or semantic flexibility demonstrated by Mistral models in handling paraphrased queries. However, LLaMA models maintain a commendable balance between accuracy and inference speed, making them an attractive option for scenarios where computational resources are limited. However, like the Phi-2 and Mistral models, they struggle with extracting specific types of information, such as full datetime values, only retrieving partial details.

    \textbf{Impact of Quantization and Model Size}. Quantization plays a crucial role in optimizing model performance based on available resources. For example, llama-2-7b-chat, mistral-7b-instruct-v0.2, and mistral-7b-openorca demonstrate that higher-bit quantization tends to improve response quality, with higher ROUGE scores, but at the cost of increased RAM usage and inference time. Results from these three models indicate that for applications demanding precise text generation, Q3, Q4, and Q5 quantizations are preferable.

    \textbf{Resource Efficiency Analysis}. A comparative analysis of memory and computational efficiency reveals that some models, like llama-2-7b-chat, manage to maintain high exact match scores (76.0\%) with a low memory footprint (2.95 GB storage and 5.45 GB RAM in the Q3\_K\_S configuration). This makes llama-2-7b-chat ideal for resource-limited environments, though it comes with a trade-off in speed, with inference times ranging from 76.92 ms to 153.51 ms. The dolphin-2\_6-phi-2.Q5\_K\_M model demonstrates the potential for applications requiring low latency, such as real-time services, with a fast inference time of 25.72 ms. However, with an exact match of 70.0\%, it performs less accurately than other open-source models selected for this study.

    \textbf{Implications for Real-World Applications}. The performance of various language models highlights their suitability for different real-world machine reading comprehension (MCR) tasks, depending on the application’s needs for accuracy, inference speed, computational resources, and financial considerations.

    \begin{itemize}
    \item 
    GPT-3 models, including GPT-3.5 and GPT-4, excel in tasks that demand deep understanding and nuanced comprehension. However, their use can be costly, as API access to these models may incur significant expenses, especially for large-scale applications. Additionally, GPT-3 models are better suited for non-sensitive use cases due to the centralized nature of their API and data management policies, which can present challenges in handling confidential or sensitive information.
    \item 
    Phi-2 models are optimal for applications where low latency and computational efficiency are prioritized. These models are a good choice for less complex tasks or scenarios with limited computational resources, offering a more efficient alternative when system performance and resource constraints are a key consideration.
    \item 
    Mistral models provide a balanced solution for applications needing both performance and speed, such as interactive systems or customer support.
    \item 
    Llama models manage to maintain high exact match scores (76.0\%) with a low memory footprint (2.95 GB storage and 5.45 GB RAM in the Q3\_K\_S configuration). This makes them ideal for resource-limited environments.
    \end{itemize}

    \begin{table*}[ht]
        \centering
        \resizebox{\linewidth}{!}{
        \addtolength{\tabcolsep}{-0.37em}
        \begin{tabular}{|l|c|c|c|c|c|c|c|}
        \hline
        \textbf{Models} & \makecell{\textbf{Size} \\ \textbf{ (GB) $\downarrow$}} &  \makecell{\textbf{Required}\\ \textbf{RAM (GB) $\downarrow$}}  & \makecell{\textbf{Inference}\\ \textbf{Time (ms) $\downarrow$}}
        & \multicolumn{1}{l|}{\makecell{\textbf{Exact Match} \\ \textbf{(\%) $\uparrow$}}}
        & \multicolumn{1}{l|}{\makecell{\textbf{Rouge-2} \\ \textbf{(\%) $\uparrow$}}}
        & \multicolumn{1}{l|}{\makecell{\textbf{Rouge-L} \\ \textbf{(\%) $\downarrow$}}}        \\ \hline
        GPT-4 & $\times$ & $\times$ &  13.21 & \textbf{87.0} & \textbf{83.0} & \textbf{94.0}\\ 
        
        GPT-3.5 & $\times$ & $\times$ &  \textbf{5.61} & 83.00 & 80.00 &  91.00\\ 
        \hline
        llama-2-7b-chat.Q2\_K & 2.83 & 5.33 & 119.65 & 71.00 &  71.00 & 84.00\\
        llama-2-7b-chat.Q3\_K\_S & 2.95 & 5.45 & 153.51 & 76.00 &  76.00 & 88.00\\
        llama-2-7b-chat.Q3\_K\_M & 3.30 & 5.80 & 76.92 &  72.00 & 71.00 &  84.00 \\

        llama-2-7b-chat.Q3\_K\_L & 3.60 & 6.10 & 91.46 & 74.00 &  72.00 & 86.00\\
        llama-2-7b-chat.Q4\_K\_S & 3.86 & 6.36 & 82.63 & 72.00 &  71.00 & 84.00\\
        llama-2-7b-chat.Q4\_K\_M & 4.08 & 6.58 & 90.34 & 72.00 &  71.00 & 85.00\\
        llama-2-7b-chat.Q5\_K\_M & 4.78 & 7.28 & 125.81 & 72.00 &  72.00 & 86.00\\
        \hline
        dolphin-2\_6-phi-2.Q5\_K\_M & \textbf{2.07} &	\textbf{4.57} & 25.72 & 70.00 & 75.00 & 87.00\\
        
        dolphin-2\_6-phi-2.Q6\_K & 2.29	& 4.79 & 28.17 & 70.00 & 74.0 & 85.0\\
        
        dolphin-2\_6-phi-2.Q8\_0 & 2.96 &	5.46 & 35.37 & 69.00 & 74.00 & 86.0\\ 
        \hline 
        mistral-7b-instruct-v0.2.Q2\_K& 3.08 & 5.58 & 38.11 & 70.00 & 73.00 & 83.00\\ 
        
        mistral-7b-instruct-v0.2.Q3\_K\_S & 3.16 & 5.66 & 42.6 & 70.00& 73.00 & 83.00\\ 
        
        mistral-7b-instruct-v0.2.Q3\_K\_M& 3.52 	& 6.02  & 42.61 & 76.00 & 75.00 & 86.00  \\ 
        
        mistral-7b-instruct-v0.2.Q3\_K\_L& 3.82 & 6.32 & 46.79 & 75.00 & 73.00 & 86.00\\ 
        
        mistral-7b-instruct-v0.2.Q4\_K\_S &  4.14 & 6.64 & 49.75 & 76.00 & 77.00 & 88.00\\ 
        
        mistral-7b-instruct-v0.2.Q4\_K\_M& 4.37 	& 6.87 & 49.75 & 76.00 & 77.00 & 88.00\\ 
        
        mistral-7b-instruct-v0.2.Q5\_K\_M & 5.13 & 7.63 & 65.15 & 74.00 & 75.00 & 88.00\\

        \hline
        
        mistral-7b-openorca.Q2\_K &  3.08 	& 5.58 & 91.84 & 75.00 & 77.00 & 88.00\\ %
        mistral-7b-openorca.Q3\_K\_S & 3.16 	& 5.66 & 84.53 & 80.00 & 80.00 & 91.00\\ %
        mistral-7b-openorca.Q3\_K\_M & 3.52 	& 6.02 & 42.31 & 83.00 & 80.00 & 90.00\\ %
        mistral-7b-openorca.Q3\_K\_L & 3.82 	& 6.32 & 115.28 & 81.00 & 80.00 & 91.00\\ %
        mistral-7b-openorca.Q4\_K\_S & 4.14 	& 6.64 & 78.20 & 81.00 & 79.00 & 89.00\\ %
        mistral-7b-openorca.Q4\_K\_M & 4.37 	& 6.87 & 80.37 & 80.00 & 78.00 & 88.00\\ %
        mistral-7b-openorca.Q5\_K\_M & 5.13	& 7.63 & 155.80 & 82.00 & 79.0 & 90.0\\ %
        \hline
        dolphin-2.6-mistral-7b.Q5\_K\_M & 5.13 & 7.63 & 63.84 & 83.00 & 79.00 & 90.00\\
        
        dolphin-2.6-mistral-7b.Q6\_K & 5.94 & 8.44 & 76.89 & 82.00 & 81.00 & 92.00 \\
        
        dolphin-2.6-mistral-7b.Q8\_0 & 7.70 & 10.20 & 94.70 & 82.00 & 81.00 & 92.00\\
        \hline
        \end{tabular}
       }
        \caption{Evaluation on our dataset. $Q_j$: quantization using $j$ bit width, K: the use of k-means clustering in the quantization, $S$, $M$, $L$: Small, Medium, and Large model size after quantization. Best results of each category are in bold.}
        \label{tab:results_models_ours}
    \end{table*}

\section{Conclusions and Future Directions}
\label{sec:Conclusions}
        This paper compares the performance and efficiency of proprietary models, such as GPT-4 and GPT-3.5, against recent open-source alternatives, like Mistral and LLaMA-2 models, for MRC tasks. The results show that proprietary models deliver fast and accurate state-of-the-art performance. GPT-4, in particular, achieved an exact match score of 87.0\% and a ROUGE-2 score of 83.0\%, making it particularly well suited for applications requiring fast and high-quality responses. However, due to concerns over the confidentiality of the information being processed, its use remains limited to contexts where data is not sensitive. Recent advances in data security, such as integrating models into secure environments via solutions like Azure OpenAI, open new perspectives for the use of proprietary models in regulated industries. These solutions provide additional guarantees for data protection, with features like data encryption and multi-factor authentication, ensuring greater compliance with security standards

    Open-source models, while showing slightly lower scores in terms of precision and speed, offer an interesting alternative due to their deployment flexibility. For example, Mistral-7b-OpenOrca achieved an 83\% exact match and a ROUGE-2 score of 80\%, while LLaMA-2 showed a 76\% exact match, proving their competitiveness in controlled and secure environments. These open-source models, with their optimized attention mechanisms and adjusted quantization configurations, show that they can compete with proprietary models while allowing companies to customize the models according to their specific needs. These models represent viable and cost-effective solutions for sectors where data privacy and the ability to deploy on private infrastructures are essential.
    
    Furthermore, the ongoing development of open-source models and the adaptation of instruct fine-tuning techniques to meet the specific needs of various industries suggest a growing adoption of these technologies across different industrial environments. These advancements could not only improve the performance of open-source models in MRC tasks but also make them more competitive with proprietary solutions, particularly in sectors where flexibility, customization, and local data management are priorities.




\newpage
\textbf{Acknowledgments}

We would like to acknowledge Novelis for their support in publishing this article. We are especially grateful for the assistance and contributions of their research team.



\bibliographystyle{elsarticle-harv} 
\bibliography{thebibliography}

\newpage

\end{document}